\pdfoutput=1 
\documentclass[letterpaper,english,american,reprint,  longbibliography, superscriptaddress]{revtex4-1}
\usepackage[T1]{fontenc}
\usepackage[latin9]{inputenc}
\usepackage{babel}
\usepackage{booktabs}
\usepackage{amsmath}
\usepackage{amssymb}
\usepackage{graphicx}
\usepackage[unicode=true]
 {hyperref}

\makeatletter

\newcommand{\noun}[1]{\textsc{#1}}
\providecommand{\tabularnewline}{\\}

\makeatother

\begin{document}
\title{Renormalized Mutual Information for Artificial Scientific Discovery}
\author{\selectlanguage{english}%
Leopoldo Sarra}
\email{leopoldo.sarra@mpl.mpg.de}

\affiliation{\selectlanguage{english}%
Max Planck Institute for the Science of Light, Erlangen, Germany}
\affiliation{\selectlanguage{english}%
Department of Physics, Friedrich-Alexander Universit�t Erlangen-N�rnberg,
Germany}
\author{\selectlanguage{english}%
Andrea Aiello}
\affiliation{\selectlanguage{english}%
Max Planck Institute for the Science of Light, Erlangen, Germany}
\author{\selectlanguage{english}%
Florian Marquardt}
\affiliation{\selectlanguage{english}%
Max Planck Institute for the Science of Light, Erlangen, Germany}
\affiliation{\selectlanguage{english}%
Department of Physics, Friedrich-Alexander Universit�t Erlangen-N�rnberg,
Germany}
\date{\today}
\begin{abstract}
We derive a well-defined renormalized version of mutual information
that allows to estimate the dependence between continuous random variables
in the important case when one is deterministically dependent on the
other. This is the situation relevant for feature extraction, where
the goal is to produce a low-dimensional effective description of
a high-dimensional system. Our approach enables the discovery of collective
variables in physical systems, thus adding to the toolbox of artificial
scientific discovery, while also aiding the analysis of information
flow in artificial neural networks.
\end{abstract}
\pacs{}
\keywords{}

\maketitle

\emph{Introduction}. \textendash{} One of the most useful general
concepts in the analysis of physical systems is the notion of collective
coordinates. In many cases, ranging from statistical physics to hydrodynamics,
the description of a complex many-particle system can be dramatically
simplified by considering only a few collective variables like the
center of mass, an order parameter, a flow field, or vortex positions.
However, in new situations, it is not clear a priori which low-dimensional
``feature'' $y=f(x)$ is best suited as a compact description of
the high-dimensional data $x$. This is the domain of unsupervised
feature extraction in computer science, where large datasets like
images or time series are to be analyzed \citep{bengio_representation_2013}.
Future frameworks of artificial scientific discovery \citep{king_automation_2009,schmidt_distilling_2009,wu_toward_2019,iten_discovering_2020}
will have to rely on general approaches like this, adding to the rapidly
developing toolbox of machine learning for physics \citep{dunjko_machine_2017,mehta_high-bias_2019,carleo_machine_2019}.

\begin{figure}
\includegraphics[width=0.8\columnwidth]{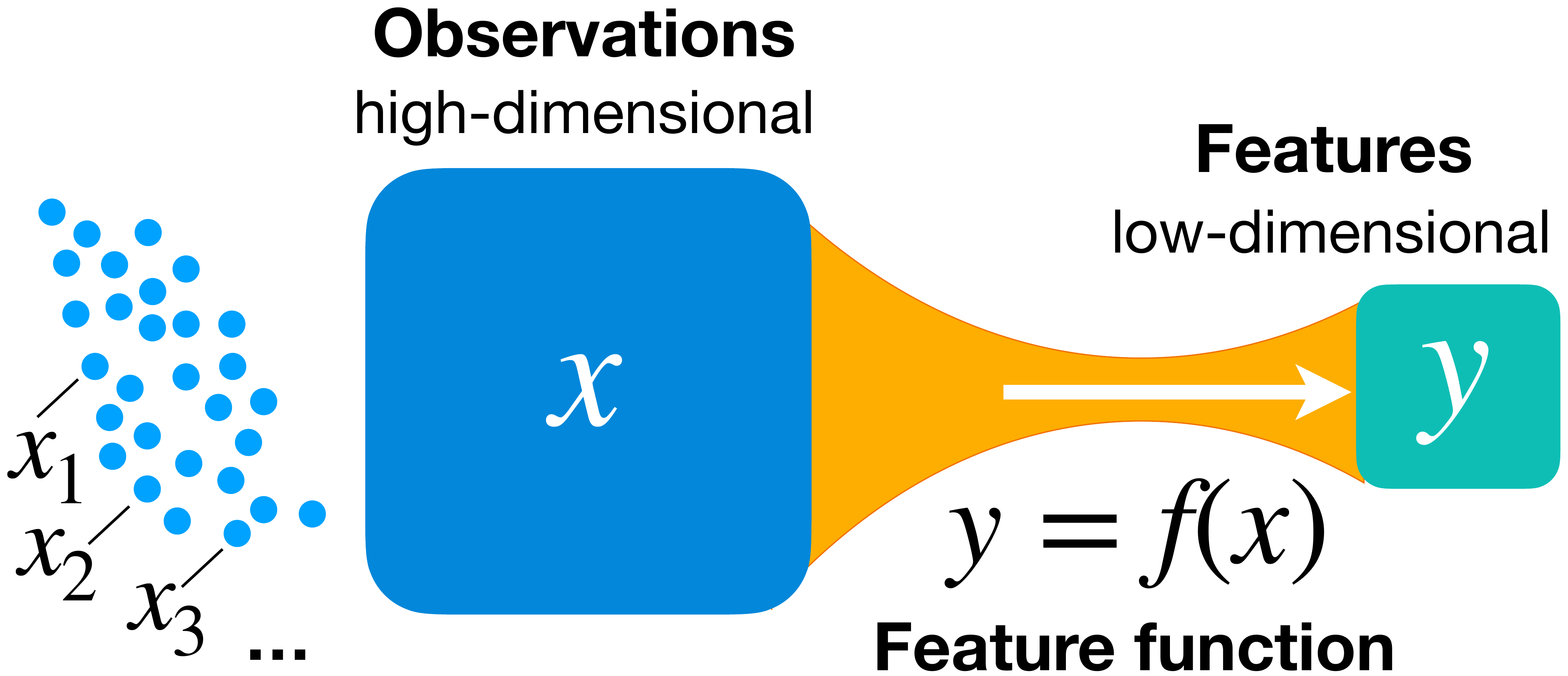}\caption{\label{fig:sketch}Feature extraction, where a high-dimensional \textquotedbl microscopic\textquotedbl{}
description $x$ (such as the configuration of a many-particle system)
is mapped to a low-dimensional feature $y=f(x)$. This is the case
where the renormalized mutual information presented in this article
is needed for feature optimization.}
\end{figure}

The simplest and most known algorithm to obtain such features is the
Principal Component Analysis (PCA) \citep{jolliffe_principal_2016}.
The idea is to project the input into the directions of largest variance.
However, its power is limited, since it can only extract linear features.
A general approach to estimate the quality of a proposed feature is
given by Mutual Information \citep{cover_elements_2006,papoulis_probability_2009}.
In general, the mutual information $I(x,y)$ answers the following
question: if two random variables $y$ and $x$ are dependent on one
another, and we are provided with the value of $y$, how much do we
learn about $x$? Technically, it is defined via $I(x,y)=I(y,x)=H(y)-H(y|x)$,
where $H(y|x)$ is the conditional entropy of $y$ given $x$ \citep{papoulis_probability_2009}.
Maximization of mutual information can be used to extract ``optimal''
features \citep{bell_information-maximization_1995}, as sketched
in Fig. \ref{fig:sketch}. 

There exists, however, a well-known important problem in evaluating
the mutual information for \emph{continuous} variables with a \emph{deterministic}
dependence \citep{amjad_learning_2019,kolchinsky_caveats_2019}, which
is exactly the case relevant for feature extraction. In this case,
$I(x,y)$ diverges, and it is not clear how to properly cure this
divergence without losing important properties of $I$. Specifically,
reparametrization invariance turns out to be crucial: applying a bijective
function to obtain $y'=g(y)$ does not change the information content,
and thus $I(x,y')=I(x,y)$.

In this work, we introduce a properly \emph{renormalized }version\emph{
}of mutual information for the important case of feature extraction
with continuous variables:
\begin{eqnarray}
\tilde{I}(x,y)=H(y)-\int dxP_{x}(x)\ln\sqrt{\det\nabla f(x)\cdot\nabla f(x)} & \quad\label{eq:mi-def}
\end{eqnarray}
where $x\in\mathbb{R}^{N}$, $y=f(x)\in\mathbb{R}^{K}$; we use $\nabla f(x)\cdot\nabla f(x)$
as a short-hand notation for $\left(\sum_{i}\partial_{i}f_{\mu}\partial_{i}f_{\nu}\right)_{\mu\nu}$,
with $1\leq i\le N$ and $1\leq\mu,\nu\leq K$, i.e. the $K\times K$
matrix resulting from the product of the ($K\times N$) Jacobian matrix
$\nabla f(x)$ and its transpose. The quantity $\tilde{I}$ is well-defined
and finite. In addition, it preserves fundamental properties of mutual
information \textendash{} among which the invariance under reparametrization
of the features:

\begin{equation}
\tilde{I}(x,g(y))=\tilde{I}(x,y),\label{eq:reparam}
\end{equation}
for a bijective function $g:\mathbb{R}^{K}\to\mathbb{R}^{K}$. We
will derive and discuss below the meaning and usefulness of the renormalized
quantity $\tilde{I}$.

Mutual information is used in many cutting edge machine learning applications,
helping to improve the intermediate layers of a neural network \citep{hjelm_learning_2018,tschannen_mutual_2020-1},
to increase the interpretability of Generative Adversarial Networks
\citep{chen_infogan_2016}, to analyze the behavior of neural networks
during training \citep{shwartz-ziv_opening_2017,saxe_information_2019}
through the Information Bottleneck method \citep{tishby_information_1999,gabrie_entropy_2019},
and for feature extraction via mutual information optimization \citep{haykin_neural_1999}.
It can be also used to characterize the variables in a renormalization
group procedure \citep{koch-janusz_mutual_2018}. Its practical estimation
is not trivial \citep{kraskov_estimating_2004}, but recently derived
bounds \citep{poole_variational_2019} permit its evaluation even
in high-dimensional spaces, with the help of neural networks \citep{belghazi_mutual_2018}.

However, there is a problem with deterministically-dependent continuous
features: the conditional entropy $H(y|x)$ formally diverges as $-\log\delta(0)$
whenever $y$ is a deterministic function of $x$. To understand why,
it is enough to take its definition, $H(y|x)=-\int dxdyP_{x}(x)P(y|x)\ln P(y|x)$,
and plug in $P(y|x)=\delta(y-f(x))$. This is specific to continuous
variables: with discrete variables, conditional entropy would be zero
and mutual information would coincide with the entropy of one of the
variables. It is clear that, to deal with a deterministic continuous
dependence, it is necessary to somehow redefine mutual information.
Past remedies involved adding noise to the feature $y$ or (equivalently)
to simply consider the non-diverging term $H(y)$ \citep{deco_information-theoretic_1996,haykin_neural_1999},
as briefly suggested in the InfoMax seminal paper \citep{bell_information-maximization_1995}.
However, they all lead to a very undesireable property: they break
the fundamental reparametrization invariance of mutual information.
In this scheme, any two features can be made to have the same entropy
$H(y)$ simply by rescaling. Thus, in the context of feature optimization,
they would be considered equally favorable, even if they represent
very different information about $x$. The reason is that such a scheme
completely ignores the diverging quantity $H(y|x)$. In contrast,
we show that $H(y|x)$ contains a non-trivial finite dependence on
the feature $f(x)$, which must be taken into account to obtain consistent
results. 

\emph{Renormalized Mutual Information}. \textendash{} In any physical
system, there are small pre-existing measurement uncertainties associated
with extracting the microscopic observables $x$. Thus, loosely speaking,
when trying to deduce information about $x$ given the value of $y$,
we have to be content with resolving $x$ up to some spread $\varepsilon$.
Motivated by this, we first consider a finite regularized quantity
$I_{\varepsilon}(x,y).$ It is defined as the mutual information between
the observable $x$ and the feature function applied to a noisy version
of the observable: $y=f(x+\varepsilon\lambda)$, where $\varepsilon\in\mathbb{R}$
is the noise strength and $\lambda\in\mathbb{R}^{N}$ is a random
multidimensional Gaussian of zero mean and unit covariance matrix.
In the limit $\varepsilon\to0$ we recover the original definition
of mutual information, which diverges logarithmically. Even in that
limit, the nature of the adopted noise distribution (e.g. isotropy,
independence of $x$) still matters, and corresponds to imposing some
hypotheses about the observed quantities $x$ (e.g. same measurement
uncertainty in all variables). We discuss these generalizations at
the end of this work. 

Consider 
\begin{eqnarray}
P(y|x)=\int d\lambda P_{\lambda}(\lambda)\delta(y-f(x+\varepsilon\lambda)).\label{eq:cond-prob}
\end{eqnarray}
When $\varepsilon\ll1$, we can expand $f(x+\varepsilon\lambda)\simeq f(x)+\varepsilon\lambda\cdot\nabla f(x)$.
By explicit calculation, it can be easily found that $P(y|x)$ is
a Gaussian distribution of zero mean and covariance matrix $\varepsilon^{2}\nabla f(x)\cdot\nabla f(x)=\varepsilon^{2}\left(\sum_{i}\partial_{i}f_{\mu}\partial_{i}f_{\nu}\right)_{\mu\nu}$.
We can calculate the conditional entropy and get
\begin{eqnarray}
H(y|x)=\int dxP_{x}(x)\ln\sqrt{\det\nabla f(x)\cdot\nabla f(x)}+KH_{\varepsilon}, & \quad\label{eq:cond-entrop}
\end{eqnarray}
where $H_{\varepsilon}$ is the entropy of a one-dimensional Gaussian
with variance $\varepsilon^{2}$. The first term only depends on the
features, and the second only on the noise. Only this term diverges
when $\varepsilon\to0$. Therefore
\begin{eqnarray}
\tilde{I}_{\varepsilon}(x,y)=I_{\varepsilon}(x,y)+KH_{\varepsilon}\label{eq:reg-noisy-i}
\end{eqnarray}
has a well defined limit $\varepsilon\to0$ and still contains all
the dependence on $f(x)$. By performing the limit we obtain our main
result, Eq.~\eqref{eq:mi-def}.

We can easily show that Eq.~\eqref{eq:mi-def} is invariant under
feature reparametrization. Consider an invertible function $z=g(y):\mathbb{R}^{K}\to\mathbb{R}^{K}$.
We can rewrite the entropy of $z$ as the entropy of $y$ plus an
extra term, which cancels with that obtained by differentiating $\ln\det(\nabla g(f(x)))$,
leading to Eq.~\eqref{eq:reparam}. We emphasize the importance of
this property: after an invertible transformation on the variable
$y$, no information should be lost, and the new variable should have
the same mutual information with $x$ as the old one. In contrast,
by adding Gaussian noise $\eta$ to the feature $y$ instead of to
$x$, i.e. $y=f(x)+\varepsilon\eta$, the final result would depend
on the feature only via $H(y)$. Reparametrization invariance would
not hold anymore under this alternative regularization: we have $I_{\varepsilon}(x,g(f(x)+\varepsilon\eta))=I_{\varepsilon}(x,f(x)+\varepsilon\eta)$
but not $I_{\varepsilon}(x,g(f(x))+\varepsilon\eta)=I_{\varepsilon}(x,f(x)+\varepsilon\eta)$
as Eq.~\eqref{eq:reparam} would require.

The price for a finite mutual information between two deterministically-dependent
variables is that when there is no dependence, e.g. $y=\text{const.}$,
we get $-\infty$ instead of $0$. In addition, given the different
roles that $x$ and $y$ play, renormalized mutual information is
no longer symmetric in its arguments. From a different perspective
\footnote{See Supplemental Material at {[}URL will be inserted by publisher{]}
for the equation of Renormalized Mutual Information in terms of information
loss.}, Eq.~\eqref{eq:mi-def} can be expressed as a particular kind of
\emph{Information Loss} \citep{geiger_information_2011,geiger_information_2011-1}. 

Mutual information obeys inequalities like $I(x,(y_{1},y_{2}))\geq I(x,y_{1})$,
which translate to the regularized version $I_{\varepsilon}$. However,
naively taking $\varepsilon\rightarrow0$ results in an empty inequality
$\tilde{I}(x,(y_{1},y_{2}))+\infty\geq\tilde{I}(x,y_{1})$. By contrast,
starting from $I(x,(y_{1},y_{2}))\geq I(x,y_{1})+I(x,y_{2})-I(y_{1},y_{2})$,
we can take the same limit and obtain a useful finite result: 
\begin{equation}
\tilde{I}(x,(y_{1},y_{2}))\geq\tilde{I}(x,y_{1})+\tilde{I}(x,y_{2})-I(y_{1},y_{2}).\label{eq:ineq1}
\end{equation}
In the special case where the dimensions of $y_{1}$ and $y_{2}$
add up to the dimension of $x$, and the mapping $x\mapsto(y_{1},y_{2})$
is bijective, reparametrization invariance produces $\tilde{I}(x,(y_{1},y_{2}))=\tilde{I}(x,x)=H(x)$,
and so

\begin{equation}
H(x)\geq\tilde{I}(x,y_{1})+\tilde{I}(x,y_{2})-I(y_{1},y_{2})\,.\label{eq:ineq2}
\end{equation}
If one constructs $y_{2}$ to be independent of $y_{1}$, the third
term on the right-hand side vanishes. However, it would be impermissible
to drop $\tilde{I}(x,y_{2})$, since it can have any sign.

\begin{figure}
\includegraphics[width=1\columnwidth]{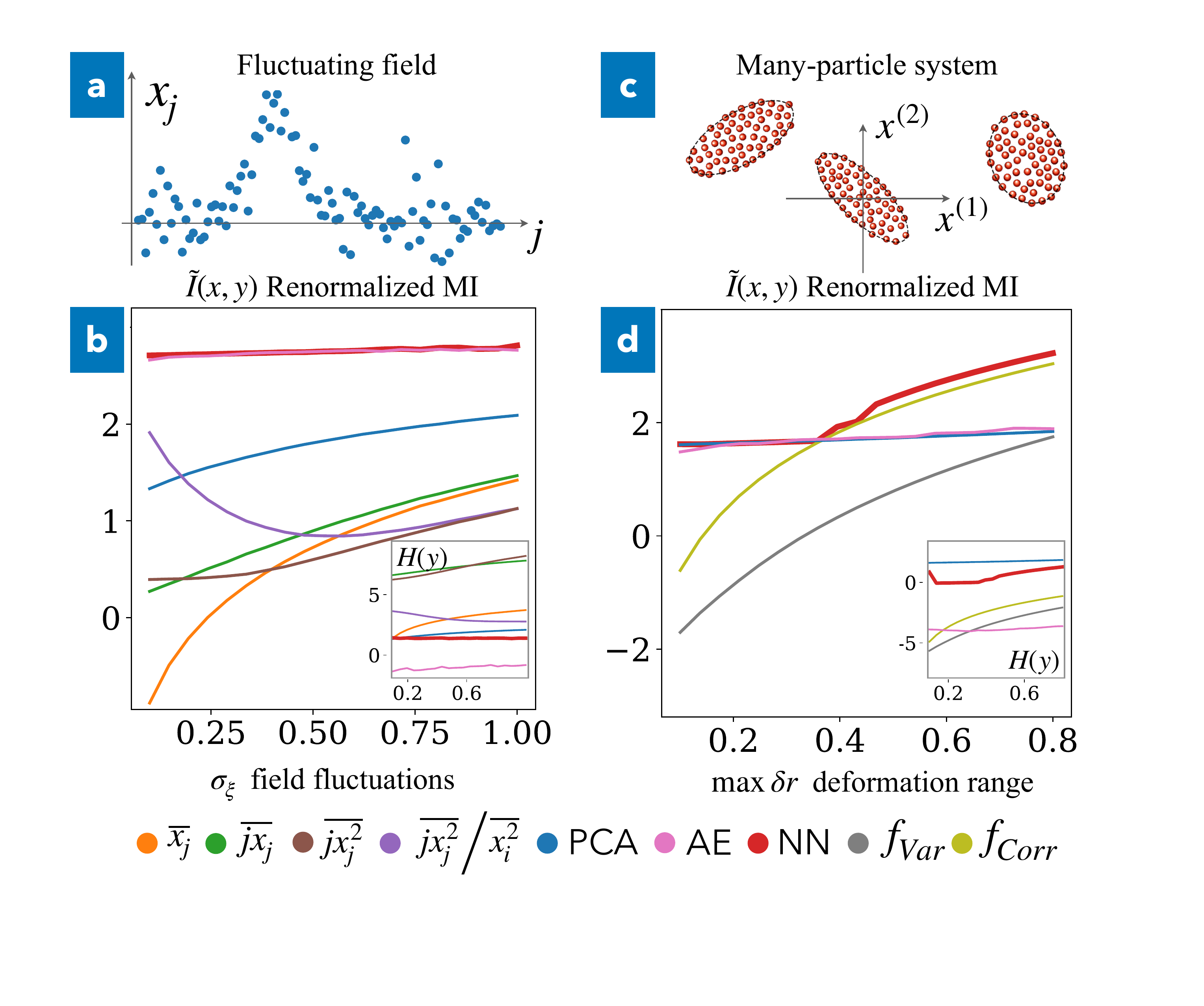}\caption{\label{fig:feature_selection} Comparing renormalized mutual information
$\tilde{I}$ for several features in two representative physical scenarios.
(a) Fluctuating 1D field on a lattice, with a randomly placed ``wave
packet'' (we depict one single sample). (b) $\tilde{I}$ as a function
of the size of the field fluctuations $\sigma_{\xi}$ for several
features. Let $\overline{A_{j}}=\frac{1}{N}\sum_{j=1}^{N}A_{j}$.
We consider: the average field $f(x)=\overline{x_{j}}$, the position
$j$ weighted by the field amplitude, $\overline{jx_{j}}$, or weighted
by the field intensity, $\overline{jx_{j}^{2}}$, as well as the ``normalized''
feature\textbf{ }$\overline{jx_{j}^{2}}/\overline{x_{i}^{2}}$ (similar
to an expectation value in quantum mechanics) and the first PCA component.
(c) Two-dimensional ``drops'' with elliptical shapes of fixed area
but with fluctuating deformation amplitude $\delta r$ and orientation
$\theta$ (we depict three samples). (d) $\tilde{I}$ vs. max. deformation
spread for the 2d-feature given by PCA and for two nonlinear features
sensitive to shape deformations, $f_{\text{Var}}=(\overline{(x_{j}^{(1)})^{2}},\overline{(x_{j}^{(2)})^{2}})$
and $f_{\text{Corr}}=(\overline{(x_{j}^{(1)})^{2}},\overline{x_{j}^{(1)}x_{j}^{(2)}})$,
where $x_{j}^{(1)},x_{j}^{(2)}$ are the coordinates of particle $j$.
In both (b) and (d): AE represents the bottleneck of a contractive
autoencoder trained to reconstruct the input and NN corresponds to
the feature given by a neural network optimized to maximize $\tilde{I}$.
In the insets, we show the entropy $H(f(x))$. This quantity is not
reparametrization invariant.}
\end{figure}
\emph{Feature comparison}. \textendash{} The renormalized mutual information
can be used to find out how useful any given ``macroscopic'' quantity
(i.e. a feature $y=f(x)$) would be in characterizing the system.
The result depends on the statistical distribution of $x$. It might
be the Boltzmann distribution in equilibrium or a distribution of
``snapshots'' of the system configuration during some arbitrary
time evolution. When control parameters such as temperature or external
fields change the distribution of $x$, the optimal feature can change.
Intuitively, observing a feature with higher $\tilde{I}$ is more
effective in narrowing down the set of underlying configurations $x$
compatible with the observed value, thus yielding more information
about the system.

We show proof-of-concept examples in the most common domains of physics
that deal with many degrees of freedom: fluctuating fields and many-particle
systems. One important goal is to discover, without prior knowledge,
that a given fluctuating field is dominated by certain localized excitations
(like solitons and vortices) and to robustly estimate their properties
(position, shape, velocity, etc.). The simplest example is a 1D field
on a lattice with a wave packet of fixed shape at a random position
(Fig.~\ref{fig:feature_selection}a,b) \footnote{See Supplemental Material at {[}URL will be inserted by publisher{]}
for technical details on the examples.}. For now, we evaluate $\tilde{I}$ for a variety of handcrafted features,
turning to feature optimization further below. Because of reparametrization
invariance (Eq.~\eqref{eq:reparam}), the scaling of any of them
is irrelevant, as is any bijective nonlinear transformation. For comparison,
we also consider PCA \citep{jolliffe_principal_2016}, which in our
context corresponds to a feature $f(x)=\sum_{j}x_{j}u_{j}$, where
$u$ is the eigenvector associated to the largest eigenvalue of the
covariance matrix $\left\langle x_{i}x_{j}\right\rangle -\left\langle x_{i}\right\rangle \left\langle x_{j}\right\rangle $,
and the bottleneck of a contractive autoencoder \citep{rifai_contractive_2011}.

In a many-particle system (molecule, star cluster, plasma, etc.),
the goal is to discover the most meaningful collective coordinates.
A simple prototypical example is a liquid drop of fluctuating shape
and orientation, made of atoms with known force fields (Fig.�\ref{fig:feature_selection}c,d).

\begin{figure}
\includegraphics[width=1\columnwidth]{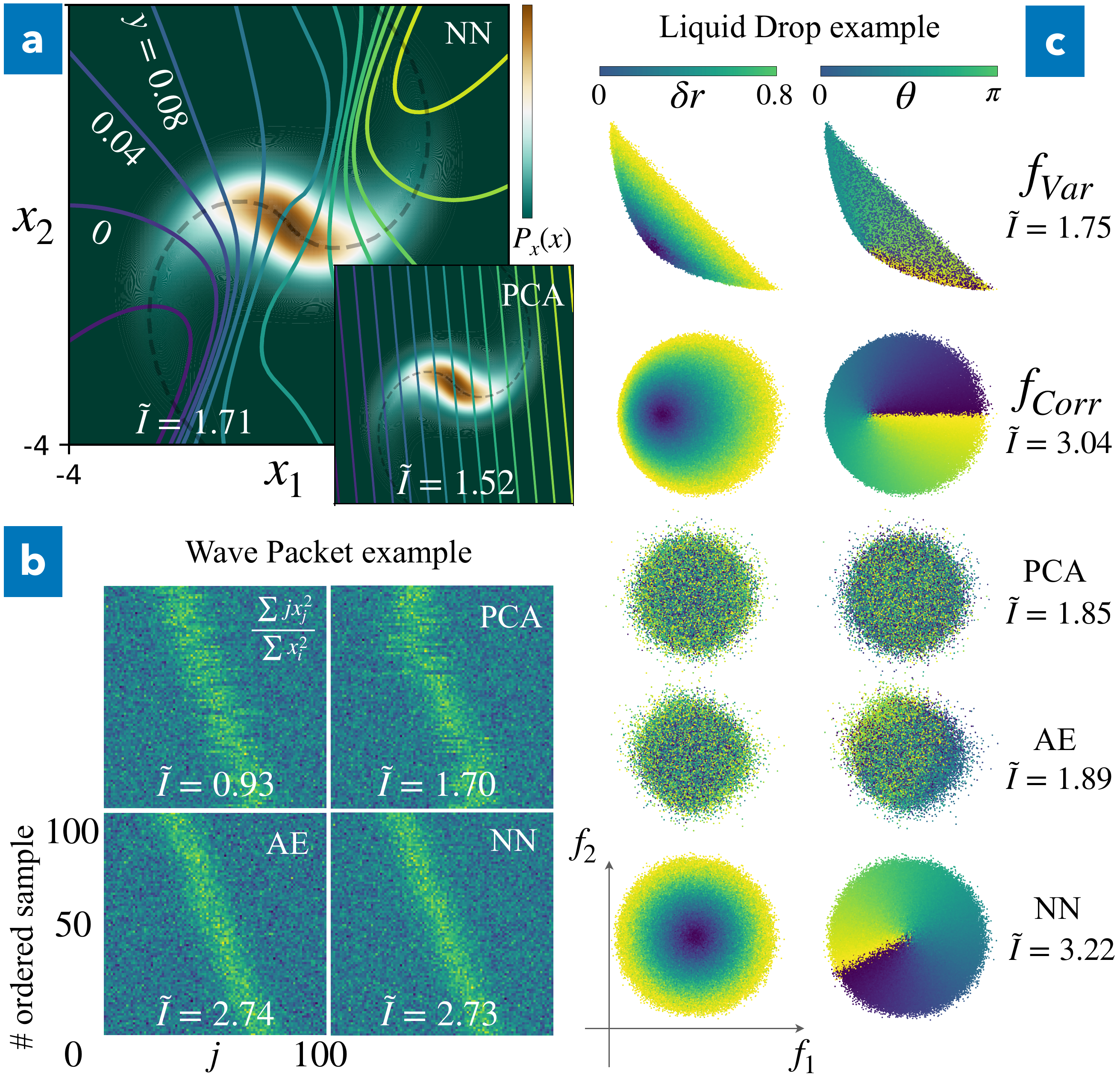}\caption{\label{fig:unconstrained}Feature optimization and visual assessment
of quality. (a) 2D non-Gaussian distribution. The obtained 1D feature
$y=f(x_{1},x_{2})$, shown as contour lines atop the distribution
$P_{x}(x)$, is parametrized with a neural network. Inset: PCA feature.
(b) Wave packets as in Fig. \ref{fig:feature_selection}a, one by
row, ordered by increasing value of the feature. The NN feature is
clearly very powerful to sort the samples. (c) Liquid drops as in
Fig. \ref{fig:feature_selection}c. We show how different 2D features
map the deformation and the orientation of the drop. The NN builds
up a representation very similar to our best handcrafted feature $f_{\text{Corr}}$.}
\end{figure}
\emph{Feature optimization}. \textendash{} Instead of comparing different
plausible features, we can consider a class of parametrized features
and optimize $\tilde{I}$ over the parameters. We opted for a multilayer
neural network \citep{goodfellow_deep_2016}, where $f(x)=f_{\theta}(x)$
with $\theta$ representing the parameters of the network. Intuitively,
meaningful features are those that provide the largest information
without over-engineering. While handcrafted features, like in the
previous section, are unarguably simple, the optimization of an excessively
powerful feature function could lead to encode additional (non-relevant)
information by means of very non-linear transformations. The tradeoff
between the simplicity of the feature and the amount of preserved
information can be adjusted both by the choice of network architecture
and by adding a small additional regularization penalty (in practice,
this can be achieved by punishing features with large gradients).
The optimization of $\tilde{I}(x,f_{\theta}(x))$ can be implemented
easily with gradient ascent algorithms \citep{goodfellow_deep_2016}.
The first term in Eq.~\eqref{eq:mi-def} can be estimated with a
histogram; for the second term, one can immediately obtain the required
$\nabla f$, since neural networks are differentiable functions, and
rely on statistical sampling of $x$. Note that also the extra degree
of freedom of feature space due to reparametrization invariance (Eq.
\ref{eq:reparam}) can be exploited to enforce additional constraints
\footnote{See Supplemental Material at {[}URL will be inserted by publisher{]}
for technical details on the numerical implementation of neural-network-based
feature optimization.}.

\begin{figure}
\includegraphics[width=1\columnwidth]{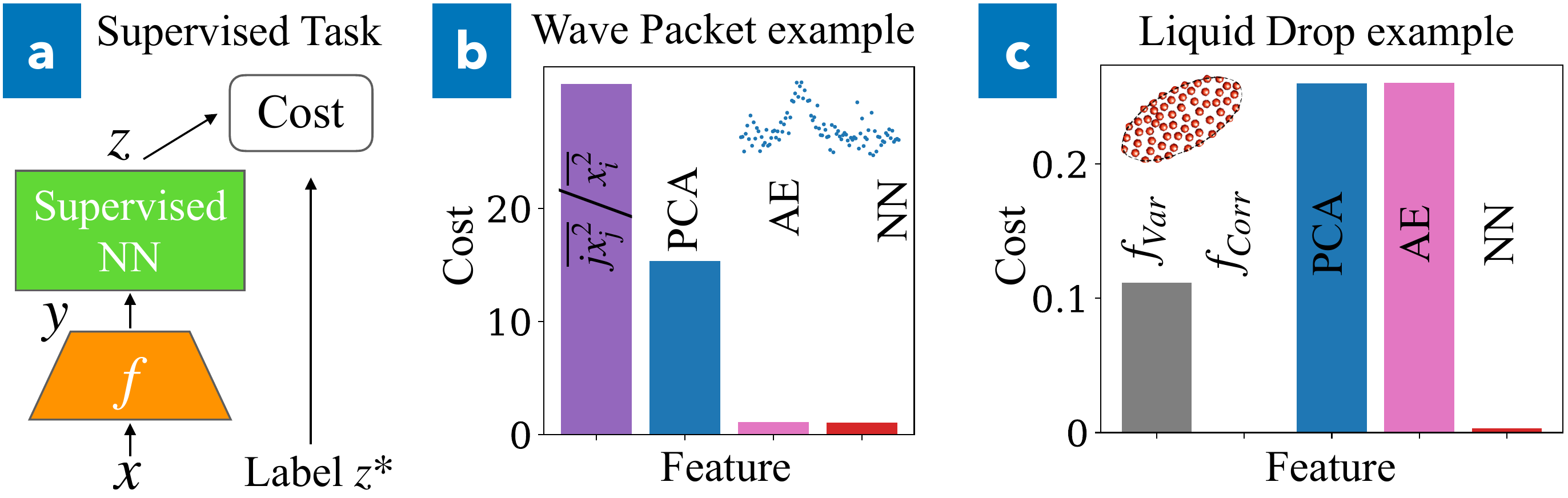}\caption{\label{fig:supervised}Comparing the performance of a supervised regression
task for different features as input. (a) For each batch of samples
$x$ we calculate the feature $y=f(x)$ and train a supervised neural
network to predict the provided label $z^{*}$. (b) Predicting the
center of the wave packet (example from Fig.~\ref{fig:feature_selection}a).
(c) Predicting the orientation and deformation of the drop (example
from Fig.~\ref{fig:feature_selection}c). The optimized NN feature
achieves the best performance in (b) and a performance very close
to that of our best handcrafted feature (c).}
\end{figure}
In Fig.~\ref{fig:unconstrained}a we show the optimization of a nonlinear
1D feature for a 2D non-Gaussian distribution. Such a low-dimensional
setting allows to visualize the shape of the feature and to compare
it with PCA. We apply the same technique also to the physical examples
(see \textquotedbl NN\textquotedbl{} in Fig. \ref{fig:feature_selection}b,
d).

One way to assess the quality of features is by suitable visualization
(see Fig.~\ref{fig:unconstrained}b,~c). The optimized NN feature
is clearly able, better than (or at least as good as) other features,
to identify the relevant properties of the system. A more quantitative,
well-known approach is to perform supervised training for a regression
task with the feature as input and analyze the resulting performance
\footnote{See Supplemental Material at {[}URL will be inserted by publisher{]}
for the technical details of the implementation.}. In the physics examples shown here, one is naturally interested
in predicting underlying parameters, like the wave packet location.
Fig.�\ref{fig:supervised}b,~c illustrate superior or very good performance
of the network.

In our illustrative examples we only considered 1D or to 2D features.
For higher-dimensional features, the numerical estimation of Eq.~\eqref{eq:mi-def}
is more challenging, but in principle still feasible \footnote{See Supplemental Material at {[}URL will be inserted by publisher{]},
section V for more details.}, for example through adversarial techniques \citep{goodfellow_generative_2014}.

Also, all the components $x_{j}$ had the same physical meaning (e.g.
particle coordinates). For components with different dimensions (e.g.
positions and momenta), one needs to decide how to compare fluctuations
along different components. A slight change in the regularization
procedure is required. Most generally, we can consider the noise distribution
$P(\lambda|x)$ to have an arbitrary covariance matrix $\Sigma(x)$,
even allowing for a location-dependent ``resolution''. We find that
it is necessary to replace the matrix $\nabla f(x)\cdot\nabla f(x)$
in Eq.~\eqref{eq:mi-def} with $\nabla f(x)\Sigma(x)\nabla f(x)$,
thus effectively introducing a metric on $x$-space \footnote{See Supplemental Material at {[}URL will be inserted by publisher{]}
for the derivation of the general case with position-dependent noise.}. This changes the inequality mentioned above (Eq.~\eqref{eq:ineq2}).

\emph{Outlook}. \textendash{} Renormalized mutual information can
be useful in many areas of statistical analysis, machine learning,
and physics.

It can be directly applied in diverse physical scenarios, with many
interesting variations and extensions. In statistical physics, one
expects that different phases of matter yield different optimal features.
Moreover, one could optimize for feature \emph{fields} (order parameter
fields) by using convolutional layers in the neural network. The locations
of defects like domain walls and vortices could be discovered as relevant
features. In general, an optimized low-dimensional description of
a high-dimensional system can be used to make partial predictions
for the time evolution. In dynamical systems the renormalized mutual
information could help to discover the underlying regularities of
the system. Even in the presence of chaos, the evolution of collective
variables can be predictable (and still non-trivial) \citep{forster_hydrodynamic_2018}.
Quantum-mechanical systems could be analyzed as well, e.g. by sampling
configurations $x$ according to a many-body state, or sampling parameters
in the Hamiltonian and looking at the expectation values $x$ of a
set of commuting observables in the corresponding ground state.

Renormalized mutual information can be used to analyze deterministic
representations of a dataset. Here we illustrated the approach only
in settings with at most two-dimensional features, but it should be
feasible to efficiently evaluate $\tilde{I}$ also with high-dimensional
feature spaces. This approach could be used to study the behavior
of a neural network from an information-theoretic perspective, for
example by analyzing the renormalized mutual information between the
input and an intermediate layer of a neural network. This could be
helpful for concepts like the ``information bottleneck'' \citep{tishby_information_1999,strouse_deterministic_2017},
which is known to be affected by the problems we discussed. Moreover,
the important challenge of representation learning for high-dimensional
datasets (like images) can benefit: our optimized features are purely
defined by their information content and not by the capability to
accomplish selected tasks. Thus, they could be useful in transfer
learning scenarios, in which many classifiers are built from the same
representation. We emphasize that the method advocated here should
be especially useful when the dimensionality is so drastically reduced
that autoencoders \citep{hinton_reducing_2006,vincent_extracting_2008,rifai_contractive_2011}
would not plausibly work very well, since it would be impossible for
a decoder to produce an approximation of the input from so few latent
variables (see Fig.~\ref{fig:unconstrained}c). This is precisely
the situation important for collective variables and similar strongly
reduced descriptions.

The code of this paper is publicly available \footnote{\href{https://github.com/lsarra/rmi}{https://github.com/lsarra/rmi}}.

\emph{Acknowledgements}. \textendash{} We thank Andreas Maier for
discussions.

\bibliographystyle{rmiaps}
\bibliography{Bibliography/bibliography}

\onecolumngrid

\appendix

\section{Derivation of Renormalized Mutual Information for the general case
of position-dependent noise}

In this section, we derive the renormalized mutual information equation,
\begin{eqnarray}
\tilde{I}(x,y)=H(y)-\int dxP_{x}(x)\ln\sqrt{\det(\nabla f(x)\cdot\nabla f(x))},\label{eq:mi-def-1}
\end{eqnarray}
in the general case in which the regularizing noise also depends on
$x$. We consider the observable distribution $x\sim P_{x}(x)$, with
$x\in\mathbb{R}^{N}.$ Let $\lambda$ be the noise variable. It has
a zero-mean Gaussian distribution with covariance matrix $\Sigma(x)$.
If we have no assumptions on the observables, we can just choose $\Sigma(x)=\mathbb{I}_{N}.$
Let $\varepsilon\in\mathbb{R}$ represent the strength of the noise.
At the end of the calculation, we perform the limit $\varepsilon\to0$.
First of all, we define the feature
\[
y=f(x+\varepsilon\lambda).
\]
Its probability distribution is given by
\[
P_{y}(y)=\int dxP_{x}(x)d\lambda P_{\lambda}(\lambda|x)\delta(y-f(x+\varepsilon\lambda)).
\]
By definition, $P_{\lambda}(\lambda|x)$ is a Gaussian distribution
with zero mean. The contribution of large values of $\lambda$ in
the $\delta$-function are suppressed by the factor $P_{\lambda}(\lambda|x)$.
As a consequence, when $\varepsilon\approx0$, we can consider the
expansion of the feature function, $f(x+\varepsilon\lambda)\approx f(x)+\varepsilon\nabla f(x)\cdot\lambda$.
We employ the Fourier representation of the $\delta$-function 
\[
\delta(y)=\frac{1}{(2\pi)^{k}}\int dse^{isy}
\]
and plug in the expression of the distribution of the noise,
\[
P(\lambda|x)=\frac{1}{\sqrt{(2\pi)^{N}\det(\Sigma(x))}}e^{-\frac{1}{2}\lambda\Sigma(x)^{-1}\lambda}.
\]
We get
\[
P(y|x)=\int\frac{ds}{(2\pi)^{k}}e^{is(y-f(x))}\int\frac{d\lambda}{\sqrt{(2\pi)^{N}\det(\Sigma(x))}}e^{-\frac{1}{2}\lambda\Sigma(x)^{-1}\lambda-is\varepsilon\nabla f(x)\cdot\lambda}=\int\frac{ds}{(2\pi)^{k}}e^{-\frac{\varepsilon^{2}}{2}s(\nabla f(x)\Sigma(x)\nabla f(x))s+i(y-f(x))s}.
\]
Now, we can also perform the Gaussian integral in $s$ and get

\[
P(y|x)=\frac{1}{\sqrt{(2\pi\varepsilon)^{k}\det(\nabla f(x)\Sigma(x)\nabla f(x))}}e^{-\frac{1}{2\varepsilon^{2}}(y-f(x))(\nabla f(x)\Sigma(x)\nabla f(x))^{-1}(y-f(x))}.
\]
This is a Gaussian distribution with mean $f(x)$ and covariance matrix
$\varepsilon\nabla f(x)\Sigma(x)\nabla f(x)$. By explicit calculation,
the conditional entropy $H(y|x)$ is given by
\[
H(y|x)=-\int dxdyP_{x}(x)P_{y}(y|x)\ln P(y|x)=\frac{K}{2}\ln2\pi e\varepsilon^{2}+\frac{1}{2}\int dxP_{x}(x)\ln\det(\nabla f(x)\Sigma(x)\nabla f(x)).
\]
We define 
\[
\tilde{I}(x,y)=\lim_{\varepsilon\to0}\left[H(y)-H(y|x)+KH_{\varepsilon}\right]=H(y)-\int dxP_{x}(x)\ln\sqrt{\det(\nabla f(x)\Sigma(x)\nabla f(x))},
\]
with $H_{\varepsilon}=\frac{1}{2}\ln2\pi e\varepsilon^{2}$. This
equation is more general than Eq.~\eqref{eq:mi-def-1} and reduces
to it if we consider an isotropic noise matrix, i.e. $\Sigma(x)=\mathbb{I}_{N}$.

\section{Reparametrization Invariance}

In this section, we verify that renormalized mutual information is
invariant under feature reparametrization. Consider an invertible
function $g(y):\mathbb{R}^{K}\to\mathbb{R}^{K}$ and the associated
random variable $z=g(y)$. Renormalized mutual information between
$x$ and $z$ can be expressed as

\begin{equation}
\tilde{I}(x,z)=H(z)-\int dxP_{x}(x)\ln\sqrt{\det(\nabla g(f(x))\cdot\nabla g(f(x)))}.\label{eq:app_rep}
\end{equation}
By employing the properties of differential entropy, we can rewrite
\[
H(z)=H(y)+\int dxP_{x}(x)\ln\det\left(\frac{dg}{dy}\right).
\]
The second term of Eq.~\eqref{eq:app_rep} can be expanded via the
chain rule of differentiation
\[
\nabla g(f(x))=\frac{dg}{dy}\cdot\nabla f(x)
\]
and by using the properties of the determinant
\[
\det(\nabla g(f(x))\cdot\nabla g(f(x)))=\left(\det\frac{dg}{dy}\right)^{2}\det(\nabla f(x)\cdot\nabla f(x)).
\]
By putting all together, we get

\[
\tilde{I}(x,z)=H(y)-\int dxP_{x}(x)\ln\sqrt{\det(\nabla f(x)\cdot\nabla f(x))}=\tilde{I}(x,y)
\]
as we wanted to show.

\section{Connection with Information Loss}

The concept of \emph{information loss }was introduced in a series
of interesting papers \citep{geiger_information_2011,geiger_information_2011-1},
as the difference between two mutual informations, $I(x,y)-I(x,z)$,
where the random variables $y$ and $z$ are functions of the random
variable $x$. The key point is that both $I(x,y)$ and $I(x,z)$
formally diverge, but their difference remains finite. Here we show
that our renormalized mutual information can be interpreted as an
information loss.

Indeed, the diverging mutual information $I(x,y=f(x))$ can be made
finite in at least two different ways: either by adding noise to the
input variables to obtain $I(x,y=f(x+\varepsilon\lambda_{x}))$, or
by adding noise to the output variables to get $I(x,z=f(x)+\varepsilon\lambda_{y})$.
Here we assume that both $\lambda_{x}$ and $\lambda_{y}$ are Gaussian
variables with zero mean and unit variance (with proper dimension).
A straightforward calculation shows that in the limit $\varepsilon\ll1$
we have 

\begin{equation}
I(x,f(x+\varepsilon\lambda_{x}))=H(y)-KH_{\varepsilon}-\int dxP_{x}(x)\ln\sqrt{\det(\nabla f(x)\cdot\nabla f(x))},\label{eq:appIL1}
\end{equation}

\begin{equation}
I(x,f(x)+\varepsilon\lambda_{y})=H(y)-KH_{\varepsilon},\label{eq:appIL2}
\end{equation}
with $H_{\varepsilon}=\frac{1}{2}\ln2\pi e\varepsilon^{2}$.

By subtracting the second equation from the first one and adding $H(y)$,
we see that the divergent term $KH_{\varepsilon}$ cancels out, and
that we obtain a relation between our \emph{finite} renormalized mutual
information and the information loss: 

\begin{equation}
\tilde{I}(x,f(x))=H(y)+\lim_{\varepsilon\to0}\left[I(x,f(x+\varepsilon\lambda_{x}))-I(x,f(x)+\varepsilon\lambda_{y})\right].\label{eq:B3}
\end{equation}
According to \citep{geiger_information_2011,geiger_information_2011-1}
the limit above represents the information lost by changing the description
of $x$ from $f(x+\varepsilon\lambda_{x})$ to $f(x)+\varepsilon\lambda_{y}$.

\section{Examples}

In this section, we give some technical details about the examples
shown in the main text. The two specific examples that we described
are illustrative of two of the most important domains of physics that
deal with many degrees of freedom: fluctuating fields and many-particle
systems. They are conceptually as simple as possible, requiring no
specialized prior knowledge.

\subsection{Spiral-shaped distribution\label{subsec:Spiral-shaped-distribution}}

The example in Fig. 3a of the main text has no particular physics
background. It is just a simple example in which unconstrained feature
optimization can be visualized directly. We consider a two-dimensional
Gaussian distribution (with zero mean) and perform the transformation
\[
\begin{cases}
x_{1}=x_{1}'\cos\alpha r'-x_{2}'\\
x_{2}=x_{1}'\sin\alpha r'+x_{2}'
\end{cases}
\]
where $r'=\sqrt{(x'_{1})^{2}+(x'_{2})^{2}}$, and $\alpha$ is a fixed
parameter. In our implementation, the covariance matrix of the initial
Gaussian is 
\[
\begin{pmatrix}0.64 & -0.56\\
-0.56 & 1
\end{pmatrix}
\]
 and $\alpha=0.5$. 

\subsection{Wave Packet\label{subsec:Wave-Packet}}

The example in Fig. 2a consists in the superposition of a fluctuating
field and a packet with a fixed shape. In particular, each sample
is given by
\begin{equation}
x_{j}=\xi_{j}+e^{-\left(j-\bar{\jmath}\right)^{2}/\delta j^{2}},\qquad j=1,\ldots,N\label{eq:wp}
\end{equation}
with $\xi_{j}$ i.i.d. Gaussian random variables, $\delta j=9$, $\bar{\jmath}$
uniformly random $\in[30,70]$, $N=100$.

\subsection{Liquid Drop\label{subsec:Liquid-Drop}}

The example in Fig.2c is built by first choosing a uniform random
deformation $\delta r\in[0,0.8]$ and orientation $\theta\in[0,\pi]$.
An ellipse with axis $A=R+\delta R$ and $B=\frac{R^{2}}{R+\delta R}$
and rotated by an angle $\theta$ is considered. The axes are chosen
so that the area of the ellipse is fixed to $\pi R^{2}$ for any $\delta r$.
$N$ particles are placed randomly inside the ellipse. Subsequently,
we turn on a Lennard-Jones interaction between the particles
\[
V(\Vert\mathbf{d}_{ij}\Vert)=\begin{cases}
-\frac{\Vert\mathbf{d}_{ij}\Vert^{2}}{d_{\text{coll}}^{2}}+\frac{3}{2}+\frac{1}{2}\left(\frac{d_{\text{eq}}}{d_{\text{coll}}}\right)^{2n}-\left(\frac{d_{\text{eq}}}{d_{\text{coll}}}\right)^{n} & \Vert\mathbf{d}_{ij}\Vert<d_{\text{coll}}\\
\frac{1}{2}\left(\frac{d_{\text{eq}}}{\Vert\mathbf{d}_{ij}\Vert}\right)^{2n}-\frac{1}{2}\left(\frac{d_{\text{eq}}}{\Vert\mathbf{d}_{ij}\Vert}\right)^{n}+\frac{1}{2} & \Vert\mathbf{d}_{ij}\Vert\geq d_{\text{coll}}
\end{cases}
\]
where $\mathbf{d}_{ij}$ is the distance between two particles, $d_{\text{coll}}$
is a cutoff at small distances and $d_{\text{eq}}$ represents the
equilibrium distance between two particles. We also add a boundary
potential that constrains the particles inside the drop
\[
V(\mathbf{r}_{i})=\begin{cases}
W\Vert\mathbf{r}_{i}\Vert & \text{outside the ellipse}\\
0 & \text{inside the ellipse}
\end{cases}
\]
where $\mathbf{r}_{i}=(x_{i}^{(1)},x_{i}^{(2)})$ is the coordinate
of a particle. We simulate the relaxation of the system by performing
some gradient descent steps; a stochastic term is added to emulate
a finite temperature $T$: we update the position of each particle
as
\[
\mathbf{r}_{i}'=\mathbf{r}_{i}+\eta\nabla_{i}V+\sqrt{2\eta T}\xi_{i},
\]
where $\xi_{i}$ is a random Gaussian variable (zero mean, unit variance)
and $\eta$ represents the step size of the thermalization. In particular,
we chose $R=1,$ $N=60,$ $n=6,$ $d_{\text{eq}}=0.27$ , $d_{\text{coll}}=0.06,$
$W=200$, $\eta=10^{-5}$. We performed $2\cdot10^{3}$ thermalization
steps.

\section{Feature Extraction in a Low-Dimensional Setting}

Here, we show how we implemented the optimization of Eq. \eqref{eq:mi-def-1}
to extract the features in the \emph{Feature optimization }section
of the main text. In particular, we extract a one- or two-dimensional
feature. In a low-dimensional feature setting, it is feasible to discretize
the feature space $y$ in a lattice, with lattice constant $\Delta y$
in the 1d case or $\Delta y_{1},\Delta y_{2}$ in the 2d case. We
keep $f(x)$ continuous and we parametrize it with a neural network,
i.e. $f(x)=f_{\theta}(x)$, where $\theta$ are the weights and the
biases of the neurons of the network. In particular, we quickly recall
that a neural network is made of many concatenated layers, and that
each layer involves the application of a non-linear function $\sigma$
(called activation function) to a linear combination of the inputs:
we can write the output of a layer of input $\mathbf{x}$ (and parameters
$\theta=(\mathbf{w},b)$) as
\[
\sigma(\mathbf{w}\cdot\mathbf{x}+b).
\]
To implement Eq. \eqref{eq:mi-def-1} in TensorFlow \citep{martin_abadi_tensorflow_2015},
we need to write it as a differentiable function. By looking at Eq.
\eqref{eq:mi-def-1}, we see that the first term $H(y)$ must be approximated
in some way; the second term can be directly used. 

The easiest way to approximate $H(y)$ is to first estimate $P_{y}(y)$,
and then use it to compute the sum $H(y)=-\Delta y\sum_{k}P_{y}(y_{k})\log P_{y}(y_{k})$
(where $\Delta y=\Delta y_{1}\Delta y_{2}$ in the 2d feature case).
We approximate the probability density $P_{y}(y)$ with a Kernel Density
Estimation procedure \citep{hastie_elements_2001}: we apply a kernel
$K_{j}(y)$ on each point $y_{j}=f(x_{j})$ and write
\[
P_{y}(y)=\frac{1}{N}\sum_{j}K_{j}(y),
\]
where $N$ is the number of points in the batch. Now we can discretize
$P_{y}(y)$ by assigning to each element of the lattice $y_{k}$
\begin{equation}
P_{y}(y_{k})=\int_{y_{k}-\frac{\Delta y}{2}}^{y_{k}+\frac{\Delta y}{2}}dyP_{y}(y),\qquad k=1,\ldots,k_{f}.\label{eq:discrPy}
\end{equation}
In other words, instead of assigning each point to a single bin of
the histogram, this function assigns it to all the bins, in a way
proportional to the kernel applied to the point. In this way, we can
always calculate analytically the result of the integral in Eq. \eqref{eq:discrPy}
as the difference of two Error functions. In practice, we use a Gaussian
kernel with variance $(s\Delta y)^{2},$i.e.
\[
K_{j}(y)=\frac{1}{\sqrt{(2\pi s^{2}\Delta y^{2})^{d}}}e^{-\left(\frac{y-f(x_{j})^{2}}{s\Delta y}\right)^{2}}
\]
(where $d$ is the dimension of $y$) and we empirically chose $s=1$
in the 1D case and $s=2$ in the 2D case. In the 1D case we discretize
the feature in $k_{f}=180$ bins, in the 2D example we use $k_{f}=100$.
We fix the bounds of the histogram so that it always includes all
the points.

The optimization of Eq. \eqref{eq:mi-def-1} is performed through
gradient descent in the following way. We define the cost function
that we want to minimize, $C=-\tilde{I}(x,f_{\theta}(x))$. At each
step, we consider a batch of samples, calculate $f_{\theta}(x)$ for
all the points in the batch and use it to estimate $P_{y}(y)$. We
calculate $\tilde{I}(x,y=f_{\theta}(x))$ and use backpropagation
to update each parameter of the neural network, i.e. an update rule
\[
\theta_{n+1}=\theta_{n}-\eta\frac{\partial C}{\partial\theta},
\]
where $n$ is the training step and the learning rate $\eta$ is a
fixed parameter of the algorithm, trying to converge to the minimum
of the cost function. In practice, one can obtain better performance
by using more advanced algorithms like RMSprop or Adam \citep{kingma_adam_2015}. 

\begin{figure}
\includegraphics[height=4cm]{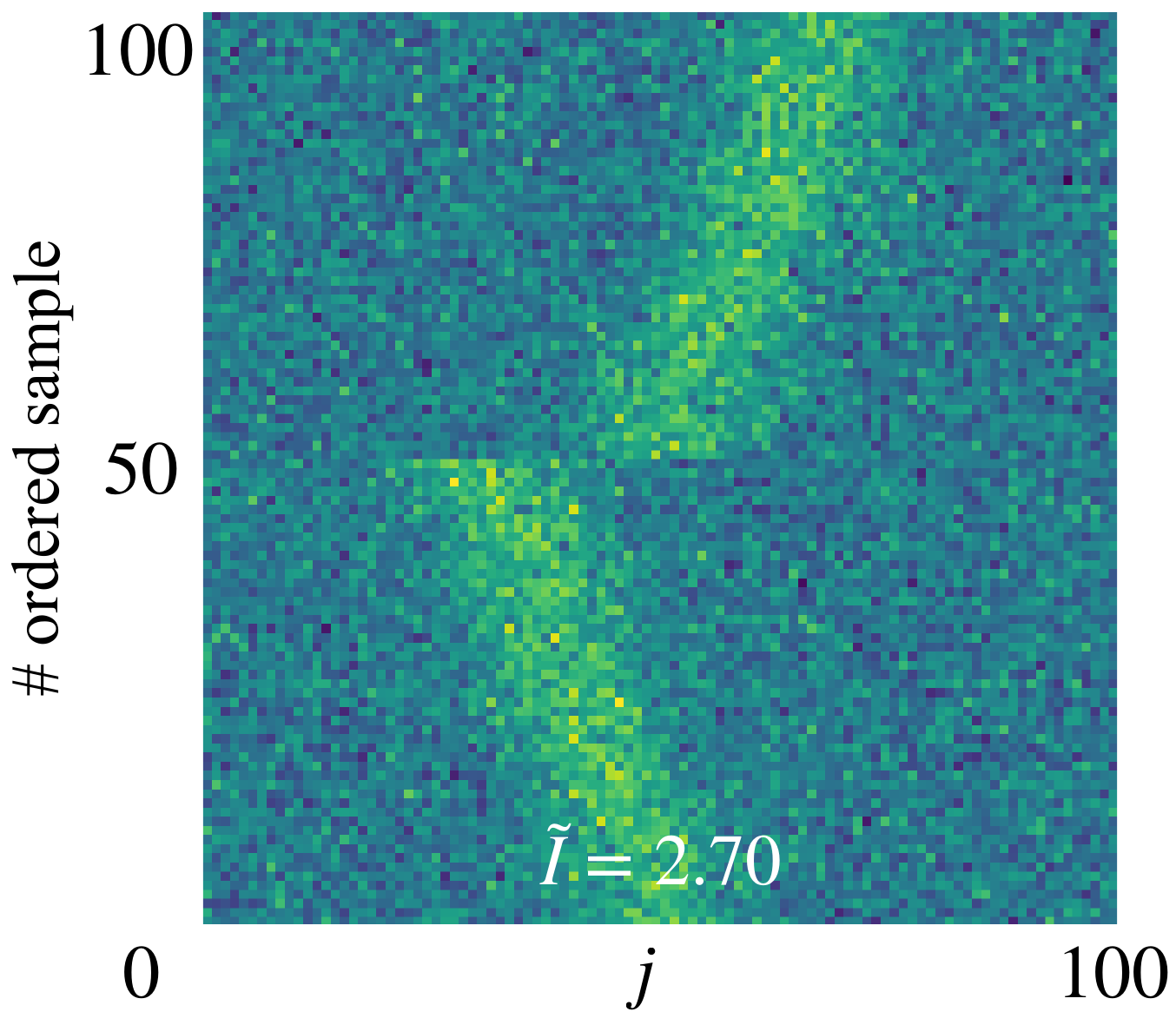}

\caption{\label{fig:wp-broken}Non-smooth feature in the case of the example
in section \ref{subsec:Wave-Packet}. We show wave packets as in Fig.
2a of the main text, one by row, ordered by increasing value of the
feature. Without the regularizing term in Eq. \ref{eq:regterm} the
optimization can get stuck in local minima that have almost the same
$\tilde{I}$ as the optimal feature (compare with Fig. 3b of the main
text). As a consequence, the network can assign different feature
values to very similar samples, while still being able to distinguish
them.}
\end{figure}

We can improve the smoothness of the extracted feature by adding to
the cost function an exponentially decaying term that penalizes large
gradients: 
\begin{equation}
Ae^{-\frac{n}{\tau}}\langle||\nabla f_{\theta}(x)||\rangle_{x}\label{eq:regterm}
\end{equation}
where $\langle a(x)\rangle_{x}=\int dxP_{x}(x)a(x)$, and $A$ and
$\tau$ are hyperparameters that should be chosen conveniently. This
can prevent extracting features that almost have the optimal information
content but present discontinuities (see for example Fig. \ref{fig:wp-broken}).

In addition, we remind that because of reparametrization invariance
(Eq. \eqref{eq:app_rep}) we can arbitrarily choose the density distribution
of the output feature. For example, we can enforce a Gaussian feature
distribution (with zero mean and variance $\sigma^{2}$) by adding
to the cost function the Kullback-Leibler divergence \citep{goodfellow_deep_2016}
of a Gaussian with the feature distribution:
\begin{equation}
B\left(KL(P_{y}||\mathcal{G})\right)=B\left(\int dyP_{y}(y)\log\frac{P_{y}(y)}{\mathcal{G}(y)}\right)=B\left(-H(y)-\frac{1}{2\sigma^{2}}\langle y^{2}\rangle_{y\sim P_{y}(y)}\right).\label{eq:gaussterm}
\end{equation}
where $B$ is a hyperparameter that can be chosen freely. This increases
the accuracy of the feature entropy because it prevents the output
distribution to condense in very small regions, making our estimate
unreliable. To sum up, the cost function that we minimize, including
all possible regularizations that we described, reads as
\begin{equation}
C=-\tilde{I}(x,f_{\theta}(x))+Ae^{-\frac{n}{\tau}}\langle||\nabla f_{\theta}(x)||\rangle_{x}+B\left(KL(P_{y}||\mathcal{G})\right).\label{eq:rmi_final_cost}
\end{equation}
Table \ref{tab:rmi-extraction} shows the parameters we used in the
examples described in Fig. 3 of the main text.

\subsection*{Improving feature extraction}

In this paper, we optimized renormalized mutual information to extract
only a one- or two-dimensional feature. As shown in the previous section,
we could use a histogram-like approximation of the entropy term in
Eq. (\ref{eq:mi-def-1}). Generalizations to high-dimensional feature
settings will clearly require to find a better way to estimate \emph{$H(y)$.}
One more advanced option to estimate it is to exploit adversarial
techniques. One could introduce a reference distribution $P_{r}(y)$
and rewrite $H(y)=-\langle\ln P_{r}(y)\rangle_{y\sim P_{y}(y)}-KL(P_{y}||P_{r})$,
where the last term is the Kullback-Leibler divergence \citep{goodfellow_deep_2016}
between $P_{r}(y)$ and $P_{y}(y)$. This term can be estimated by
means of adversarial techniques \citep{goodfellow_generative_2014},
i.e. by optimizing over an auxiliary ``discriminator'' neural network
$D(y)$: 
\begin{align*}
H(y) & =-\langle\ln P_{r}(y)\rangle_{y\sim P_{y}(y)}+\min_{D}\left(\langle D(y)\rangle_{y\sim P_{y}(y)}+\langle e^{-D(y)}\rangle_{y\sim P_{r}(y)}-1\right).
\end{align*}
For convenience, $P_{r}(y)$ can be chosen as a Gaussian distribution.
This is related to the technique proposed in \citep{belghazi_mutual_2018}
to estimate mutual information, while properly allowing for the renormalization
discussed here.

\begin{table}
\noindent\begin{minipage}[c]{1\columnwidth}%
\begin{tabular}{ccc}
\toprule 
\multicolumn{3}{c}{\textbf{Spiral-shaped distribution}}\tabularnewline
\midrule 
Layer & Neurons & Activation\tabularnewline
\midrule
\midrule 
input & $2$ & -\tabularnewline
\midrule 
layer 1 & $30$ & tanh\tabularnewline
\midrule 
layer 2 & $30$ & tanh\tabularnewline
\midrule 
output & $1$ & linear\tabularnewline
\midrule
\midrule 
 &  & \tabularnewline
\midrule 
cost & $\tilde{I}$ & \tabularnewline
\midrule 
optimizer & Adam & \tabularnewline
\midrule 
learning rate & $5\cdot10^{-3}$ & \tabularnewline
\midrule 
batch size & $100$ & \tabularnewline
\midrule 
training steps & $3\cdot10^{4}$ & \tabularnewline
\midrule 
regularization: $A$ & 0 & (see Eq. \ref{eq:regterm})\tabularnewline
\midrule 
regularization: $B$ & $5$ & (see Eq. \ref{eq:gaussterm})\tabularnewline
\bottomrule
\end{tabular}\hfill{} %
\begin{tabular}{ccc}
\toprule 
\multicolumn{3}{c}{\textbf{Wave packet}}\tabularnewline
\midrule 
Layer & Neurons & Activation\tabularnewline
\midrule
\midrule 
input & $100$ & -\tabularnewline
\midrule 
layer 1 & $70$ & tanh\tabularnewline
\midrule 
layer 2 & $70$ & tanh\tabularnewline
\midrule 
output & $1$ & linear\tabularnewline
\midrule
\midrule 
 &  & \tabularnewline
\midrule 
cost & $\tilde{I}$ & \tabularnewline
\midrule 
optimizer & Adam & \tabularnewline
\midrule 
learning rate & $5\cdot10^{-3}$ & \tabularnewline
\midrule 
batch size & $700$ & \tabularnewline
\midrule 
training steps & $1.5\cdot10^{4}$ & \tabularnewline
\midrule 
regularization: $A$ & $10^{2}$ & (see Eq. \ref{eq:regterm})\tabularnewline
\midrule 
regularization: $\tau$ & $10^{3}$ & (see Eq. \ref{eq:regterm})\tabularnewline
\midrule 
regularization: $B$ & $5$ & (see Eq. \ref{eq:gaussterm})\tabularnewline
\bottomrule
\end{tabular} \hfill{}%
\begin{tabular}{ccc}
\toprule 
\multicolumn{3}{c}{\textbf{Liquid drop}}\tabularnewline
\midrule 
Layer & Neurons & Activation\tabularnewline
\midrule
\midrule 
input & $120$ & -\tabularnewline
\midrule 
layer 1 & $800$ & relu\tabularnewline
\midrule 
output & $2$ & linear\tabularnewline
\midrule
\midrule 
 &  & \tabularnewline
\midrule 
cost & $\tilde{I}$ & \tabularnewline
\midrule 
optimizer & RMSprop & \tabularnewline
\midrule 
learning rate & $5\cdot10^{-3}$ & \tabularnewline
\midrule 
batch size & $5\cdot10^{3}$ & \tabularnewline
\midrule 
training steps & $3\cdot10^{4}$ & \tabularnewline
\midrule 
regularization: $A$ & $15$ & (see Eq. \ref{eq:regterm})\tabularnewline
\midrule 
regularization: $\tau$ & $500$ & (see Eq. \ref{eq:regterm})\tabularnewline
\midrule 
regularization: $B$ & $5\cdot10^{-2}$ & (see Eq. \ref{eq:gaussterm})\tabularnewline
\bottomrule
\end{tabular}%
\end{minipage}\caption{\label{tab:rmi-extraction} Layout and training parameters of the
network used for feature optimization. (left) Spiral-shaped distribution
as in Section \ref{subsec:Spiral-shaped-distribution}. (middle) Wave
packet as in Section \ref{subsec:Wave-Packet}. (right) Liquid drop
as in Section \ref{subsec:Liquid-Drop}. When we used the gradient
regularization as in Eq. \ref{eq:regterm}, we always carried the
optimization until the term completely decayed and was irrelevant
in the final part of the training.}
\end{table}

\section{Autoencoders}

Autoencoders \citep{goodfellow_deep_2016} are a particular state-of-the-art
kind of neural network intended for representation learning in unsupervised
settings. They have two parts: an Encoder, which takes the input $x$
and outputs a lower dimensional representation $f(x)$, the feature,
also called latent space or bottleneck, and the Decoder, which takes
the latent representation $f(x)$ and whose output has the same dimension
of the input. This neural network is optimized so that the output
matches the input. Since the bottleneck has a lower dimension than
the input and output, only an insightful representation will allow
to optimally reconstruct the input. We used a particular kind of autoencoder
called Contractive Autoencoder \citep{rifai_contractive_2011}, which
should encourage more robust features. Apart from minimizing the mean
squared error between the input and the reconstructed output, it also
includes a term that penalises the Frobenius norm of the Jacobian
of the bottleneck: 
\[
\lambda||\nabla f(x)||^{2}.
\]
In our experiments, we consider the feature provided by the output
of the encoder (i.e. the function that maps the input to the bottleneck),
after the autoencoder has been trained. We used the same networks
described in Table \ref{tab:rmi-extraction} for the encoder and an
analogous network for the decoder (but we used relu activation functions
in the decoders). The training parameters (optimizer, learning rate,
batchsize and number of training steps) are the same as in Table \ref{tab:rmi-extraction}.
We used $\lambda=10^{-2}.$

\section{Feature Performance in a Supervised Task}

To obtain a more quantitative comparison with alternative unsupervised
feature extraction techniques, we have adopted the conventional approach
for comparing the performance of representations: we carry out supervised
training of a neural network that takes as its input the feature.
It is clear a-priori that it cannot be possible to solve arbitrary
prediction tasks based on a very low-dimensional (1D or 2D) feature.
However, in many physical scenarios such as the ones we consider,
there are only very few underlying salient properties which might
be of interest. This motivated the choice of our tasks: to predict,
from the feature, the wavepacket center-of-mass or the deformation
and orientation of the drop, respectively.

\begin{figure}
\noindent\begin{minipage}[t][1\totalheight][c]{1\columnwidth}%
\begin{center}
\begin{tabular}{ccc}
\toprule 
\multicolumn{3}{c}{\textbf{Supervised Network - Wave Packet}}\tabularnewline
\midrule 
Layer & Neurons & Activation\tabularnewline
\midrule
\midrule 
input & $1$ & -\tabularnewline
\midrule 
layer 1 & $50$ & relu\tabularnewline
\midrule 
layer 2 & $50$ & relu\tabularnewline
\midrule 
output & $1$ & linear\tabularnewline
\midrule
\midrule 
 &  & \tabularnewline
\midrule 
cost & MSE & \tabularnewline
\midrule 
optimizer & Adam & \tabularnewline
\midrule 
learning rate & $10^{-3}$ & \tabularnewline
\midrule 
batch size & $200$ & \tabularnewline
\midrule 
training steps & $10^{4}$ & \tabularnewline
\bottomrule
\end{tabular}\hfill{}%
\begin{minipage}[c]{0.45\columnwidth}%
\begin{center}
\includegraphics[scale=0.25]{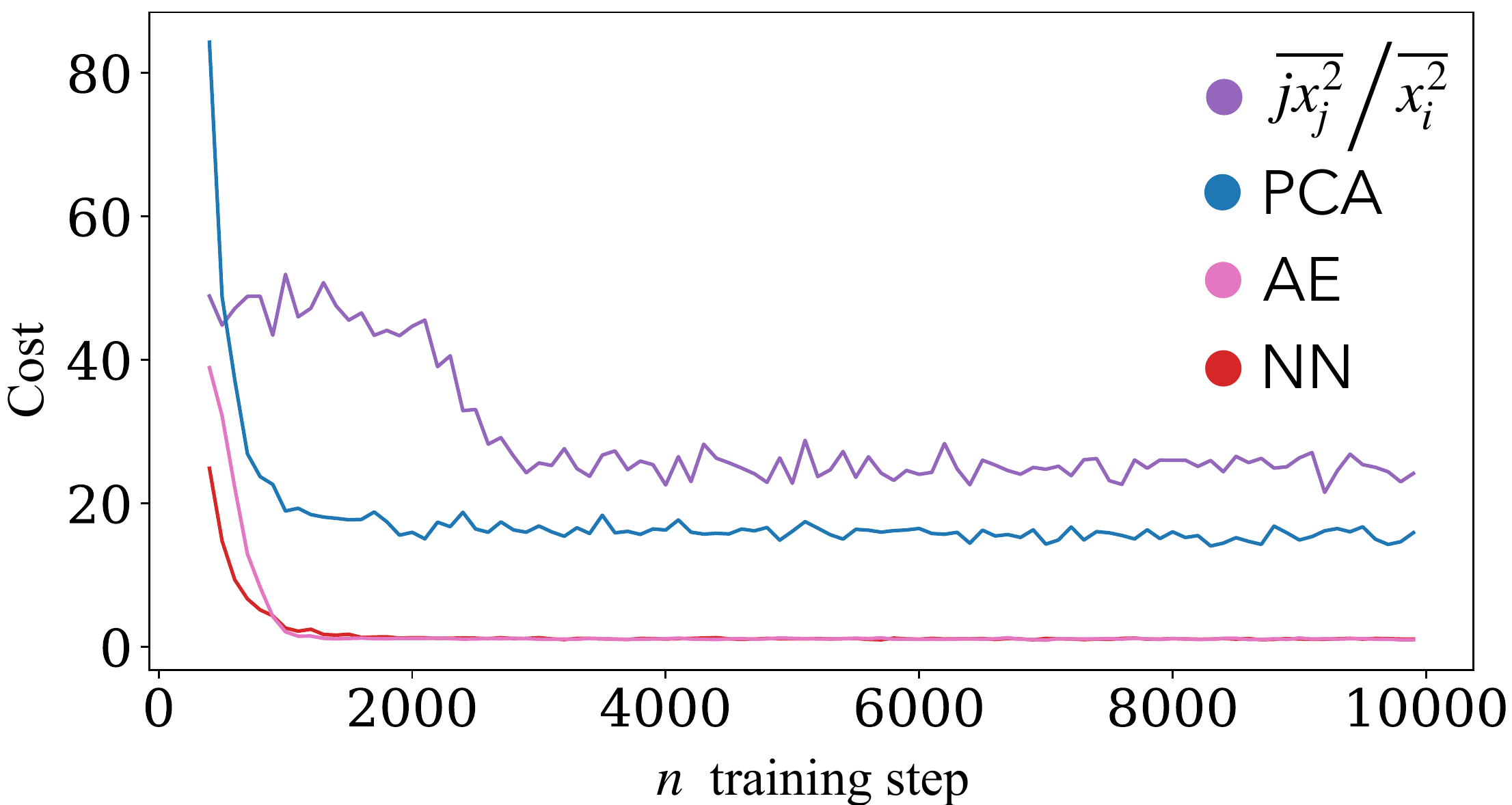}
\par\end{center}%
\end{minipage}\hfill{}%
\begin{tabular}{|c|c|c|}
\hline 
Feature & $\tilde{I}$ & Cost\tabularnewline
\hline 
\hline 
$f_{E}$ & $0.93$ & $31.0$\tabularnewline
\hline 
PCA & $1.71$ & $15.4$\tabularnewline
\hline 
AE & $2.75$ & $1.12$\tabularnewline
\hline 
NN & $2.73$ & $1.05$\tabularnewline
\hline 
\end{tabular}
\par\end{center}%
\end{minipage}\caption{\label{tab:wp-table} Performance comparison of the supervised task
in Fig. 4b of the main text. (left) Structure of the supervised neural
network and optimization parameters. (middle) For each feature, we
train a neural network whose goal is to reconstruct the center of
the wave packet. (right) Performance of a manually engineered feature
$f_{E}=\sum_{j=1}^{N}jx_{j}^{2}/\sum_{i=1}^{N}x_{i}^{2}$, the first
component given by Principal Component Analysis (PCA), the feature
obtained at the bottleneck of an autoencoder (AE) and the one given
by optimizing Renormalized Mutual Information (NN).}
\end{figure}

In Fig. 4a of the main text, we considered the example described in
Section \ref{subsec:Wave-Packet}. At fixed field fluctuation strength
$\sigma_{\xi}=0.38$, the supervised neural network should reconstruct
the center of the wave packet, given one of the features. We use the
Mean Squared Error (MSE) between the output of the network and the
value of the center of the wave packet as cost function. The results
are shown in Figure \ref{tab:wp-table}. It is evident that the feature
given by optimizing Renormalized Mutual Information allows for a better
performance. 

In the case of the example described in Section \ref{subsec:Liquid-Drop},
the supervised neural network should reconstruct the orientation and
deformation of the liquid drop. To reconstruct an angular variable
like the orientation of the liquid drop, we cannot directly apply
the mean squared error. In this case, the supervised neural network
has three outputs, which will have to predict $\delta r$, $\cos2\theta$
and $\sin2\theta$ respectively (we remind that the boundaries of
the drops look the same when rotated by $\pi$). In addition, one
should take into account that the orientation of the drop can't be
predicted if the deformation of the drop is close to zero. Therefore,
the final cost function looks like
\begin{equation}
C=(n_{1}-\delta r^{*})^{2}+\delta r^{*}\left((n_{2}-\cos2\theta^{*})^{2}+(n_{3}-\sin2\theta^{*})^{2}\right),\label{eq:lj-cost}
\end{equation}
where $(n_{1},n_{2},n_{3})$ are the outputs of the supervised network
and $\delta r^{*}$and $\theta^{*}$ the true deformation and orientation
of the sample. 
\begin{figure}
\noindent\begin{minipage}[t][1\totalheight][c]{1\columnwidth}%
\begin{tabular}{ccc}
\toprule 
\multicolumn{3}{c}{\textbf{Supervised Network - Liquid Drop}}\tabularnewline
\midrule 
Layer & Neurons & Activation\tabularnewline
\midrule
\midrule 
input & $2$ & -\tabularnewline
\midrule 
layer 1 & $100$ & relu\tabularnewline
\midrule 
layer 2 & $100$ & relu\tabularnewline
\midrule 
output & $3$ & linear\tabularnewline
\midrule
\midrule 
 &  & \tabularnewline
\midrule 
cost & modified MSE & see Eq. \ref{eq:lj-cost}\tabularnewline
\midrule 
optimizer & Adam & \tabularnewline
\midrule 
learning rate & $10^{-3}$ & \tabularnewline
\midrule 
batch size & $1500$ & \tabularnewline
\midrule 
training steps & $2\cdot10^{4}$ & \tabularnewline
\bottomrule
\end{tabular} \hfill{}%
\begin{minipage}[c]{0.45\columnwidth}%
\begin{center}
\includegraphics[scale=0.25]{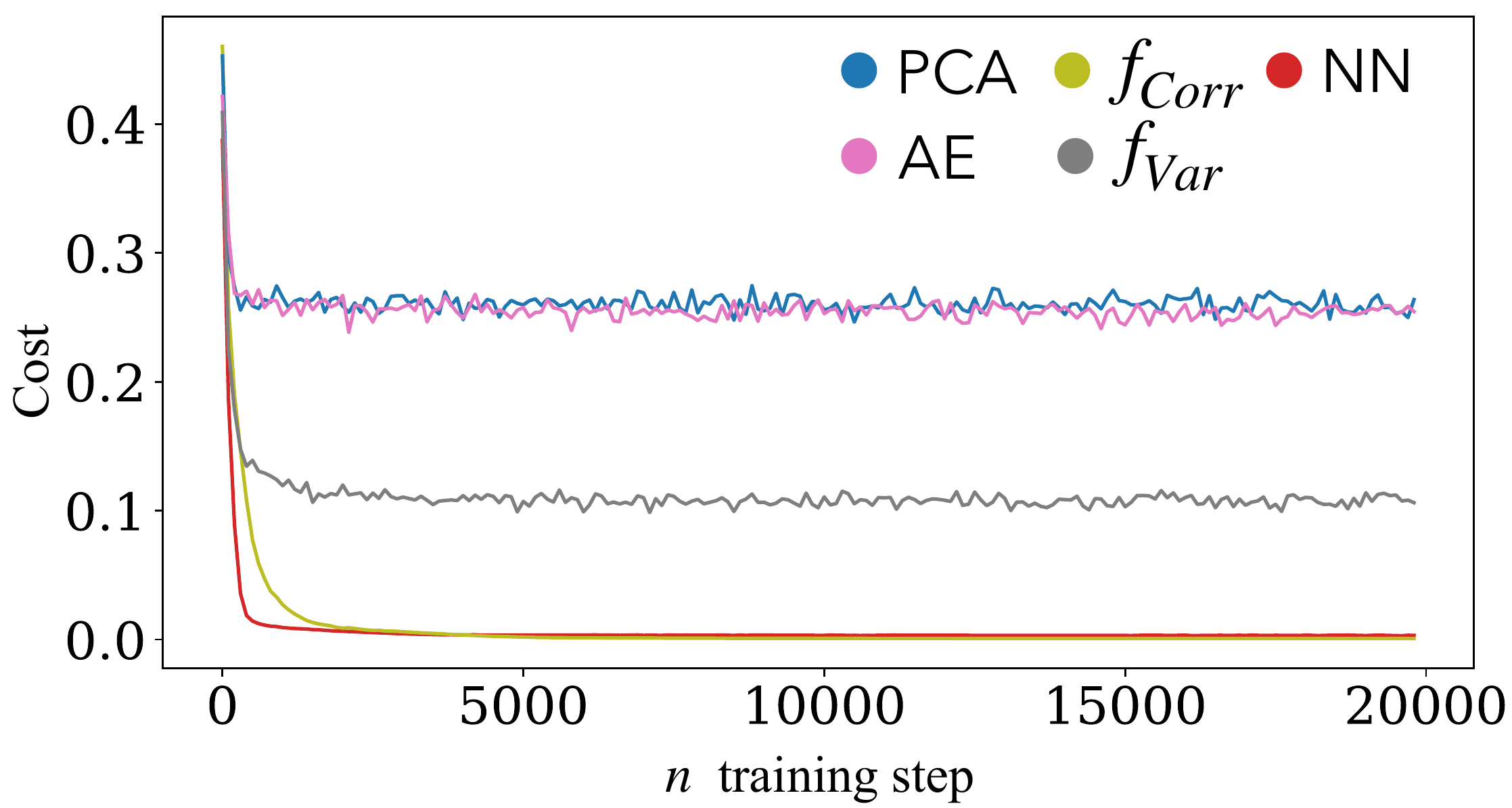}
\par\end{center}%
\end{minipage}\hfill{}%
\begin{tabular}{|c|c|c|}
\hline 
Feature & $\tilde{I}$ & Cost\tabularnewline
\hline 
\hline 
$f_{\text{Var}}$ & $1.75$ & $0.11$\tabularnewline
\hline 
$f_{\text{Corr}}$ & $3.04$ & \noun{$0.001$}\tabularnewline
\hline 
PCA & $1.85$ & $0.26$\tabularnewline
\hline 
AE & $2.18$ & $0.24$\tabularnewline
\hline 
NN & $3.21$ & $0.003$\tabularnewline
\hline 
\end{tabular}%
\end{minipage}\caption{\label{tab:1d-table} Performance comparison of the supervised task
in Fig. 4c of the main text. (left) Structure of the supervised neural
network and optimization parameters. (middle) For each feature, we
train a neural network whose goal is to reconstruct the orientation
and deformation of the drop. (right) We compare the two manually engineered
feature $f_{\text{Var}}=\frac{1}{N}\left(\sum_{j}(x_{j}^{(1)})^{2},\sum_{j}(x_{j}^{(2)})^{2}\right)$
and $f_{\text{Corr}}=\frac{1}{N}\left(\sum_{j}(x_{j}^{(1)})^{2},\sum_{j}x_{j}^{(1)}x_{j}^{(2)}\right)$,
the first component given by Principal Component Analysis (PCA), the
feature given by the bottleneck of an autoencoder (AE) and that obtained
by optimizing Renormalized Mutual Information (NN). We use the cost
function defined in Eq. \ref{eq:lj-cost}. The last column of the
table shows the final value of the cost (Eq. \ref{eq:lj-cost}) for
each of the features.}
\end{figure}
The results are shown in Figure \ref{tab:1d-table}. Again we see
that the feature obtained by Renormalized Mutual Information maximization
allows to build a supervised network that is as accurate as that built
starting from the handcrafted feature (which by design contains all
the information needed to accomplish the task).
\end{document}